\documentclass[10pt,twocolumn,letterpaper]{article}

\usepackage{iccv}
\usepackage{times}
\usepackage{epsfig}
\usepackage{graphicx}
\usepackage{amsmath}
\usepackage{amssymb}
\usepackage{graphicx}
\usepackage{amsmath}
\usepackage{amssymb}
\usepackage{booktabs}
\usepackage{wrapfig}
\usepackage{url}
\usepackage{graphicx}
\usepackage{amsthm}
\usepackage{multirow}
\usepackage{adjustbox}
\usepackage{bm}
\usepackage{bbold}
\usepackage{float}
\makeatletter
\@namedef{ver@everyshi.sty}{}
\makeatother
\usepackage{graphicx}

\usepackage{mathtools,algorithm}
\usepackage[noend]{algpseudocode}


\usepackage[breaklinks=true,bookmarks=false]{hyperref}

\iccvfinalcopy 

\def\iccvPaperID{****} 

\ificcvfinal\fi

\begin{document}

\title{Explainable Artificial Intelligence Architecture for Melanoma Diagnosis Using Indicator Localization and Self-Supervised Learning}

\author{Ruitong Sun\\
University of Southern California\\
{\tt\small ruitongs@usc.edu}
\and
Mohammad Rostami\\
University of Southern California\\{\tt\small rostamim@usc.edu}
}

\maketitle
\ificcvfinal\thispagestyle{empty}\fi

\begin{abstract}
   Melanoma is a prevalent lethal type of cancer that is treatable if diagnosed at early stages of development. Skin lesions are a typical indicator for diagnosing melanoma but they often led to delayed diagnosis due to high similarities of cancerous and benign lesions at early stages of melanoma. 
Deep learning (DL) can be used as a solution to classify skin lesion pictures with a high accuracy, but clinical adoption of deep learning faces a significant challenge. The reason is that the decision processes of deep learning models are often  uninterpretable  which makes them black boxes that are challenging to trust. We develop an explainable deep learning architecture for melanoma diagnosis which generates clinically interpretable visual explanations for its decisions. Our experiments demonstrate that our proposed architectures matches clinical explanations significantly better than existing architectures.

\end{abstract}

\section{Introduction}
\label{sec:intro}

Melanoma is a prevalent type of skin cancer that can be highly deadly in advanced stages. For this reason, early detection of melanoma is the most important factor  for successful treatment. 
New skin moles or changes in existing models are the most distinct symptoms of melanoma.
However,  due to similarity of benign and cancerous moles, melanoma diagnosis is a sensitive task that can be preformed by trained dermatologist. If skin moles are not screened and graded on time, melanoma maybe detected  by patients very late.  Unfortunately, this is the case with low-income populations with   limited access to healthcare.
Advances in deep learning along with accessibility of smartphones have led to  emergence of automatic diagnosis of melanoma using  skin lesion ordinary photographs \cite{codella2017deep,sultana2018recent,li2018skin,adegun2019deep,kassani2019comparative,naeem2020malignant,jojoa2021melanoma}. When evaluated only in terms of diagnosis of melanoma,   deep   models have accuracy rates close to those of   dermatologists. Despite this success, adoption of these models in clinical setting has been   limited.

\begin{figure}[h]
    \centering
    \includegraphics[width=6cm, height=3cm]{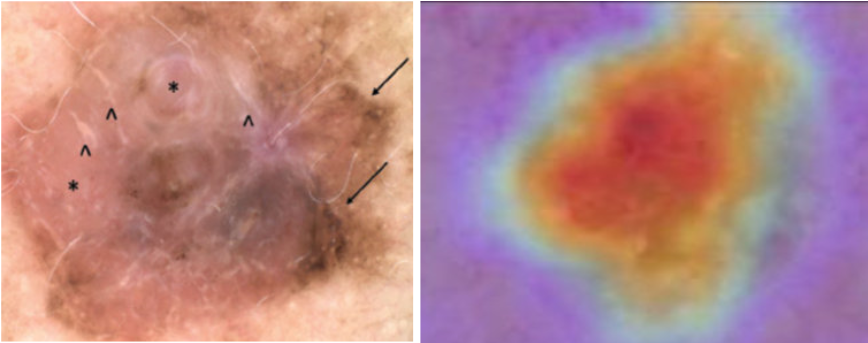}
    \caption{Left: a skin mole photograph; right: explainability heatmap generated using Grad-Cam~\cite{selvaraju2017grad} for a     trained DNN.}
    \label{fig1}
    \vspace{-5mm}
\end{figure}

A primary challenge for adopting deep learning in clinical tasks is the challenge of interpretability. Deep neural networks (DNNs) sometimes are called ``black boxes'' because their internal decision-making process is opaque. Existing explainability methods~\cite{shrikumar2017learning,selvaraju2017grad,zhang2018top,pope2019explainability} try to clarify decisions of these black boxes to help users or developers understand the most important areas of the image in making the classification, in the form of a heatmap.  However, the area alone is not particularly helpful, e.g., Grad-Cam~\cite{selvaraju2017grad} simply highlights the entire mole in the melanoma image in Figure~\ref{fig1}. In other words, the highlighted regions are often too large to show the shape of an interpretable region, or extremely deviated from the regions of interest to dermatologists. 
A reason behind this deficiency is that many explainability methods primarily consider the last DNN layer for heatmap generation, whereas some interpretable features maybe encoded at earlier layers. Hence, improving DL explanation methods may be helpful.

More importantly, there is no guarantee that a trained DNN uses human interpretable indicators for decision-making~\cite{avramidis2022automating}, irrespective of improving DL explainability algorithms.
We argue existing explainability methods may not be sufficient for explainable DL in high-stakes domain such as medicine due to the end-to-end training pipeline of DL. In other words, a model that is trained only using a high-level abstract label, e.g., cancerous vs benign, may learn to extract indicator features that are totally different compared to the features human experts use. 
In contrast, dermatologists are trained to perform their diagnosis through identifying intermediate indicator biomarkers \cite{argenziano1998epiluminescence}.
The solution that we propose is to benefit from intermediate-level annotations that denote human-interpretable features in the training pipeline to enforce a DNN learn making decisions similar to clinical experts. However, data annotation, particularly in medical applications, is an expensive and time-consuming task and generating a finely annotated dataset is infeasible. To circumvent this challenge, we   use self-supervised learning~\cite{chen2020simple} to train a  human-interpretable model using only a small annotated dataset. Our empirical experiments demonstrate that our approach can generate explanations more similar to expert dermatologists.
 
\section{Related Work}
\label{sec:related}

\paragraph{Explainability in Deep Learning}

Existing explainability methods in deep learning primarily determine which spatial regionx of the input image or combination of regions led to a specific model decision or contribute significantly to the network prediction (see Figure~\ref{fig1}). There are two main approaches to identify regions of interest when using deep learning:   Model-based methods and model agnostic methods. Model-based methods  work based on the details of the specific structures of a   deep learning model. They examine the activations or weights of the deep network to find regions of importance. Grad-CAM and Layerwise Relevance propagation \cite{samek2016evaluating} are examples of such methods. Attention-based methods \cite{dosovitskiy2020image} similarly identify important image regions.  Model agnostic methods ethods separate the explanations from the model which offers wide applicability. These methods (e.g., LIME~\cite{ribeiro2016should}) manipulate inputs (e.g., pixels, regions or superpixels) and measure how changes in input affect output. If an input perturbation has no effect, it is not relevant to decision-making. In contrast, if a change has a major impact (e.g., changing the classification from glaucoma to normal), then the region is important to the classification. SHapley Additive exPlanations (SHAP) 
    ~\cite{lundberg2017unified} can assign each feature or region an importance value for a particular prediction.  Note, however, the regions found by these algorithms do not necessarily correspond to intermediate concepts or diagnostic features that are known to experts or novices. Hence, while these algorithms are helpful to explain classifications of DNNs, they do not help training models that mimic humans when making predictions.

  Identifying regions of interest is also related to semantic segmentation \cite{noh2015learning,stan2021domain} which divides an image into segments that are semantically meaningful (e.g., separating moles from background skin in diagnosing melanoma). However, these methods mostly segment based on spatial similarities and do not offer any explanation how these segments that are generated can be used for classification of the input image.  UNets  \cite{ronneberger2015u} also identify regions within images but do not indicate the importance of regions to overall classification, a key step in explaining model decision.

\paragraph{ Deep Learning for Melanoma Diagnosis}

Dermatology is one of  the most common use cases of DL in medicine, with many existing works in the literature~\cite{codella2017deep,sultana2018recent,li2018skin,adegun2019deep,kassani2019comparative,naeem2020malignant,jojoa2021melanoma}. Despite significant progress in DL, these methods  simply train a DL on a labeled binary dataset using supervised learning. Despite being naive in terms the artificial intelligence (AI) algorithms they use, these works lead to decent performances, comparable with expert clinicians. There is still room for improving explainability characteristics of these methods to convince clinicians adopting AI for melanoma diagnosis in practice. However, only a few works have explored explainability of AI models for melanoma diagnosis. Murabayashi et al.~\cite{murabayashi2019towards} use clinical indicators benefit from virtual adversarial training~\cite{miyato2018virtual} and multitask learning to train a model that predicts the clinical indicators in addition to the binary label to improve explainability. Nigar et al.~\cite{nigar2022deep} simply use LIME to study interpretability of their algorithm.
Stieler et al.~\cite{stieler2021towards} use the ABCD-rule, a diagnostic approach of dermatologists, while training a model to improve interpretability. 
Shorfuzzaman~\cite{shorfuzzaman2022explainable} used meta-learning to train an ensemble of DNNs, each predicting an indicator, to use indicators to explain decisions. These existing works, however, do not spatially locate the indicators. We develop and architecture that generates spatial masks on the input image to locate clinical indicators spatially.


\section{Problem Formulation}
\label{sec:formulation}

Due to the privacy concerns~\cite{stansecure} and high annotation costs~\cite{rostami2018crowdsourcing} associated with medical images, available and well-annotated medical data is extremely scarce. However, there are many unannotated datasets available. Therefore, we aim to leverage these large amounts of unannotated data to improve our models. We refer to such a dataset as $D^{UL}=(\bm{x}'i)^N_{i=1}$, where $x'_i$ denotes the images. We refer to this dataset as Dataset B. 

Most works based on using AI for melanoma diagnosis, consider that we have access to a dataset that includes skin lesion images along with corresponding binary labels for cancerous vs benign cases. The standard supervised learning is then use to train a suitable binary classifier, primarily based on convolutional neural networks (CNNs). 
In our work, we also consider such a dataset is accessible, referred to Dataset A in our formulation. 
Despite being a simple procedure for AI, it has been used extensively in the   literature~\cite{premaladha2016novel,zhang2017melanoma,codella2017deep,sultana2018recent,li2018skin,adegun2019deep,kassani2019comparative,naeem2020malignant,jojoa2021melanoma} due to high accuracy rates. 
However, as explained, this simple baseline does not lead to a human-centered explainable model. We cannot benefit from unsupervised domain adaptation ~\cite{rostami2023}
because we want to transfer knowledge from the unannotated domain to the annotated domain.

Fortunately, there are a few labeled datasets available where images are annotated with clinically plausible indicators. These indicator commonly are used by dermatologists and residents are trained to diagnose melanoma based on identifying them. We try to implement a similar approach, where the model is trained in end-to-end scheme to first predict the indicators as intermediate-level labels and then use them for diagnosis label prediction.
Let $D^{L} = (\bm{x}_i, y_i,(\bm{z}_{ij})_{j=1}^d)^M_{i=1}$ denotes this dataset, where $x_i$ and $y_i$ denote the images and their binary diagnostic labels. Additionally, $z_{ij}$ denotes a feature mask array with the same size as the input image, where for each $j$, the mask  denotes the spatial location  of a clinically interpretable  indicator, e.g., pigmented network, on the input image in the form of a a binary segmentation map.
We refer to this dataset as Dataset A. Clearly, preparing Dataset A is a significantly  more challenging task that Dataset B. It suffices to go though existing medical records to prepare Dataset B according to diagnosis.
In contrast, existing medical records rarely include instances for Dataset A and hence,  a dermatologist should determine the absence and presence of each indicator and locate then on the image in addition to a simple binary label. Even if can annotate some images to generate Dataset A,  the size of Dataset A will be significantly smaller than Dataset B ($M<<N$) due to scarcity of dermatologists who would accept serving as data annotators. Our goal is to benefit from both Dataset A and Dataset B to train an architecture that can be used for melanoma diagnosis with interpretable explanations.

A naive idea to train an explainable model is to use a suitable architecture and train one segmentation model, e.g., U-Net~\cite{ronneberger2015u} to predict indicator masks. Previously, this idea has been used for training explainable AI models for medicine~\cite{sharma2022segmentation}. In our problem, we can use one U-Net for each of the $d$ indicators and train them using Dataset A. Hence, we will have $d$ image segmentation models that determine spatial location of each indicator for input images. However, there are two shortcomings. First, we will still need a secondary classification model to determine the diagnosis label from the indicators~\cite{murabayashi2019towards}. More importantly,  the size of the Dataset A may not be sufficient for this purpose, particularly because only a subset of instances will contain a particular type of indicator and semantic segmentation is a complex task compared to classification. Since we likely will encounter the challenge of attribute sparsity, we likely will face overfitting during model execution.

The idea we will explore is to benefit from information encoded in Dataset B. Although Dataset B is not attributed coarsely, it is similar to Dataset A and transferring knowledge between these two datasets might be feasible. We formulate a weakly supervised learning problem for this purpose. Specifically, we rely on self-supervised learning using Dataset B to train an encoder that can better represent input images, enabling the model to generalize and locate biomarkers. Additionally, we propose to train multiple encoders to separately learn each feature, so that we can apply unique operations to each encoder and improve performance. Therefore, we can benefit from transfer learning to address the challenge of data sparsity. Finally, we connect the output vectors of multiple encoders to predict labels $\bm{y}_i$, making the model's predictions similar to those of an expert.


\section{Proposed Algorithm}
\label{sec:algorithm}

We have developed an explainable architecture for melanoma diagnosis. The network utilizes self-supervised learning to understand the inherent data structure. Then, we use U-Net as the basic backbone, and by letting the encoder perform the downsampling task, the Resnet backbone focuses more on the important areas identified by self-supervised learning, rather than providing too general features. In this section, we will describe the network architecture components and explain why we use each part. Additionally, we will discuss the training procedure used.


\begin{figure*}[ht]
    \centering
    \includegraphics[width=.9\textwidth]{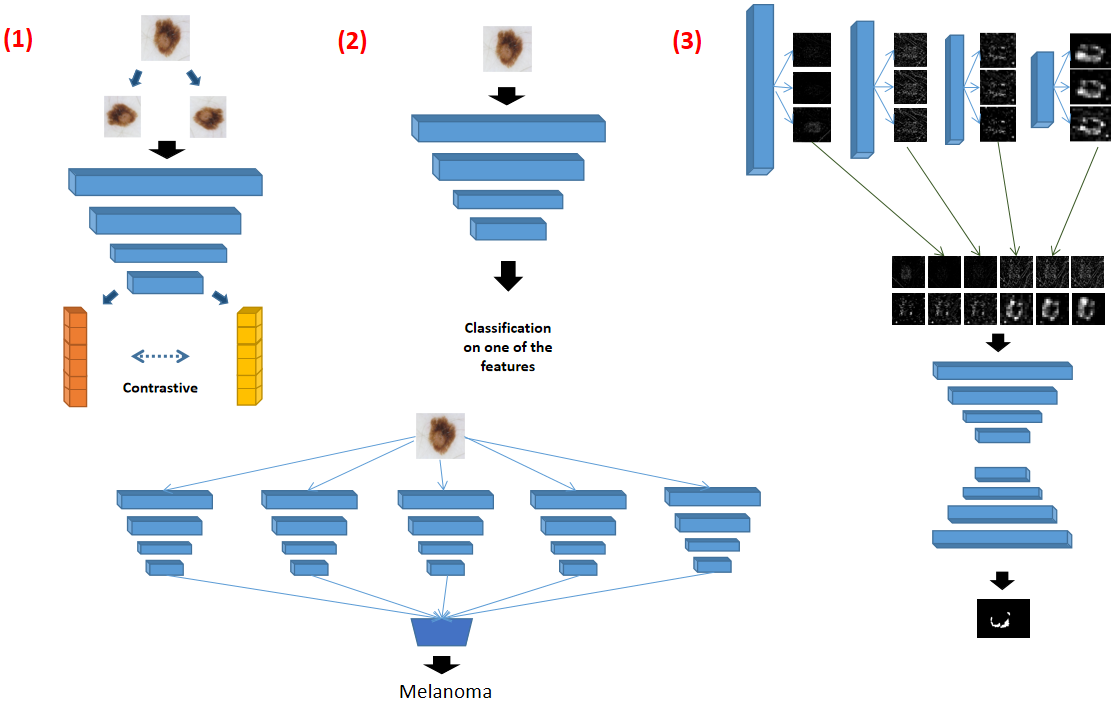}
    \caption{Proposed architecture for explainable diagnosis of melanoma: the architecture is trained to simultaneously classify skin lesion pictures using a CNN classifier and learn to localize melanoma clinical indicators on the input image using a U-Net-based segmentation network. } 
    \label{fig:P1}
\end{figure*}

\subsection{Overview of the network architecture}

 Figure \ref{fig:P1},  Visualize the architecture of our proposed model, which aims to generate a heatmap for the final convolution layer that closely resembles the regions expected by human experts when the network is trained for diagnostic classification. Our architecture consists of three key subnetworks:

 \begin{itemize}
     \item[]
 (1) A pretrained Resnet50 network for attributes classification; we denote the subnetwork (1) as $f_{Res}$ where $f_{Res}(\bm{x}_i) = E_1(\bm{x}_i) + fc_1(\bm{x}_i)$ , $\bm{x}_i \in D^L$ and $E_1(\cdot)$ represents the encoder, which consists of four consecutive blocks, each block consists of a downsampling and two ResNet residual blocks, $fc_1(\cdot)$ represents a fully connected   layer which produces probabilistic outputs.
 
 \item[] (2) An architecture based on U-Net for biomarker indicator locations. We replaced the U-Net encoder with the encoder $E_1(\cdot)$ of the $f_{Res}(\cdot)$ ; we denote the subnetwork (2) as $f_{Seg}(\cdot)$ where $f_{Seg}(\bm{h}_i) = E_1(\bm{h}_i) + D_1(\bm{h}_i)$, where $\bm{h}_i$ denotes the heatmaps that are generated using Grad-Cam, $D_1(\cdot)$ denotes the decoder which outputs biomarker indicator location.
 
 \item[] (3) We add a ResNet50-based projection head for self-supervised learning; we denote the subnetwork (3) as $f_{CLR }(\cdot)$, where $f_{CLR}(\bm{x}_i^{\prime}) = E_1(\bm{x}_i^{\prime}) + P_1(\bm{x}_i^{\prime})$ and $\bm{x}_i^{\prime} \in D^{ UL}$, $P_1(\cdot)$ is a projection head consisting of two fully connected   layers that outputs a feature vector. 
 
\end{itemize}

In summary, our architecture learns to classify the input images. Then, we use Grad-Cam to generate attention heatmaps. These heatmaps do not necessarily show interpretable indicators, but are useful for classification which means that they should have correlations with features expert clinicians use. Our idea is to feed these heatmaps to the segmentation subnetwork and generate the indicator biomarker localization maps using the Grad-Cam heatmaps. As a result, the architecture learns to generate explanations along with diagnosis labels.
We can see that our full architecture can only be trained using Dataset A.


\subsection{Bio-UNet Baseline Training}
Our Bio-Unet baseline consists of networks $f_{Res}(\cdot)$, $f_{Seg}(\cdot)$ . At first, We train the network $f_{Res}(\cdot)$ to do the skin lesion classification task using simple supervised learning. After this stage,   the classification network can predict diagnosis labels with high accuracy. We then apply   Grad-CAM on the network to generate attention heatmaps. Grad-CAM is generally used by computing the gradient of the classification score with respect to the last convolutional feature map to identify the classification score of the selected target in the most influential part of the input image. However, the final convolutional feature map primarily provides a high-level region of the input image which corresponds to the area of interest. Our experiments demonstrate that this mostly will just generate the full lesion mole similar to Figure~\ref{fig1}  which is usually much larger than the region delineated by experts for a specific indicator and usually cannot reflect the location and area of an indicator. However, the earlier convolutional feature maps contain low-level information, e.g., the boundary of a region of interest. Hence, combining attention maps at all convolutional layers looks like an option that may generate a good estimation for the location of an indicator, but we empirically observed that simply averaging all heatmaps will result in poor output. Therefore, we would like to benefit from binary masks provided by experts and  reconstruct these heatmaps using the segmentation network $f_{Seg}(\cdot)$.  We learn to use  appropriate weights for each heatmap so that all bottleneck blocks of $E_1(\cdot)$ contribute to reconstruction of the final heatmap. Because the encoder $E_1(\cdot)$ has been used for training, in order not to affect its parameters, we created an encoder $E_3(\cdot)$ with the same architecture as $E_1(\cdot)$. Specifically, the Resnet encoder $E_3(\cdot)$ consists of 12 bottleneck blocks, each loaded with optimal checkpoint parameters.Then we input a training image and a bottleneck block in Grad-CAM, get a heatmap, and then replace the bottleneck block in turn, getting a total of 12 localization masks. We denote $f_{Grad\_CAM}(b_i,j)$ where $b_i$ is the bottleneck block and j is the index of the selected label.

\indent For the segmentation task, dice coefficient is commonly used with a  value ranging from 0 to 1 as the loss function. The larger the value, the more similar two binary masks would be. However, the Dice loss is only designed for binary data. In order to avoid adding the artificial factor of threshold during the model training, We adopt the soft dice loss as the segmentation loss in Eq. \ref{eq1}. The soft dice loss directly uses the predicted probability instead of using the threshold to change the output to 0 or 1. It is defined as:
\begin{equation}
L_{Soft Dice} = 1 - \frac{2\sum_{Pixels}y_{true}y_{predict}}{\sum_{Pixels}y_{true}^2 + \sum_{Pixels}y_{pred}^2}
\label{eq1}
\end{equation}

By performing downsampling on the input heatmaps, the encoder will focus more on important features. During training, we add only the heatmap of the currently predicted feature by the encoder as input to $f_{Seg}(\cdot)$, making the classification model $f_{Res}(\cdot)$ more accurate for the current attribute's region of interest. Therefore, explanations can improve the diagnostic accuracy.

\subsection{Self-Supervised Learning for Bio-UNet}

The proposed Bio-Unet baseline architecture can learn boundaries efficiently, but this is only effective when the number of annotated images is large for biomarker indicators, e.g.,  pigment network. or the  area corresponding to the indicators on a lesion is contiguous and large. To enable classification models to generate accurate heatmaps when Dataset A is small or when the area for an indicator is very small  or is scatterred on the image, we benefit from self-supervised learning on Dataset B to improve the baseline of our proposed network architecture that is obtained by training on Dataset B. Specifically, we use SimCLR~\cite{chen2020simple} which uses contrastive learning for improved visual representation. As shown in Figure \ref{fig:P1}, there are two independent data augmenters $T_1(\cdot)$ and $T_2(\cdot)$ which are randomly selected from rotation, scaling, cropping, brightness, contrast, saturation, and flipping transforms to generate an augmented version samples of Dataset B so that we can compute the contrastive learning loss. Each training image $\bm{x}^{\prime}_i \in D^{UL}$is passed through two data augmenters to produce two augmented images. The two augmented images will then pass through our shared weight encoder $E_1(\cdot)$ and projection head $P_1(\cdot)$, resulting in two 128-length features. In a minibatch of $N$ input images, $2N$ augmented images will be produced, for each pair of augmented images we treat as positive pairs, the other $2(N-1)$ are negative examples.We adopt contrastive loss as semi-supervised loss in Eq. \eqref{eq2} 

\begin{equation}
\small
L_{CLR} = -\log\frac{\exp(sim(f_{CLR}(T_1(x_i)),f_{CLR}(T_2(x_j)))/\tau)}{\sum_{k=1}^{2N}\exp(sim(f_{CLR}(T_1(x_i)),f_{CLR}(T_2(x_k)))/\tau)},
\label{eq2}
\end{equation}
where $sim(u,v) = \frac{u^Tv}{\|u\|\|v\|}$ and k $\not=$ i and $\tau$ denotes a temperature parameter. Upon training the encoders on Dataset B using self-supervised learning, we can benefit from transferring obtained knowledge across Dataset A and Dataset B.
Due to the space limit, the full training procedure for our architecture is described in Algorithm~\ref{alg}.

\begin{algorithm}[h]
\caption{Proposed  Architecture Training Approach}\label{alg:cap}
\small
\hspace*{\algorithmicindent} \textbf{Input1:}  $(\bm{x}_i, y_i)^N_{i=1} \in D^{L}$\\ 
\hspace*{\algorithmicindent} \textbf{Input2:} $(\bm{x}'_i, y'_i)^M_{i=1} \in D^{UL}$\\
\hspace*{\algorithmicindent} \textbf{Input3:} $(b_i)^{12}_{i=1}$\\
\hspace*{\algorithmicindent} \textbf{Output1:} parameter $\theta$ for encoder $E_1$\\
\hspace*{\algorithmicindent} \textbf{Output2:} the final biomarker indicator location\\
\begin{algorithmic}[1]
\State $f_{CLR}(x_i^{\prime}) = E_1(x_i^{\prime}) + P_1(x_i^{\prime})$; 
\State $f_{Res}(x_i) = = E_1(x_i) + fc_1(x_i)$; 
\State $f_{Seg}(H) = E_1(H) + D_1(H)$ 
\State $f_{Grad\_CAM}(b_i,j)$ where $b_i \in E_3$ 
\State $T_1,T_2$ : two separate data augmentation operators\\
\While{j $<$ 5}   \Comment{For 5 attributes}
\While{Not stop}
\State Sample batch $B_1$ = $x^{\prime}_i \in D^{UL} $ 
\State Generating $f_{CLR}(T_1{(B_1)})$ and $f_{CLR}(T_2{(B_1)})$
\State Calculating loss $L_{CLR}$ as equation(2)
\State Computing gradient of $L_{CLR}$ and update $E_1$ parameters $\theta$ and $P_{1}$ parameters $\theta_1$
\State Sample batch $B_2$ = $\{(x_i,y_i) \in D^L \}$ 
\State Generating $f_{Res}(B_2)$
\State Calculating loss $L_{BCE}$
\State Computing gradient of $L_{BCE}$ and update $E_1$ parameters $\theta$ and $fc_{1}$ parameters $\theta_2$
\State Load the optimal checkpoint parameter $\theta$ on $E_3$, use $f_{Grad\_CAM}(b_i,j)$ where $b_i \in E_3$ to get the heatmap $h_i$
\State $H = (h_i)^{12}_{i=1}$
\State Generating $f_{Seg}(H)$
\State Calculating loss $L_{SoftDice}$ as equation(1)
\State Computing gradient of $L_{SoftDice}$ and update $E_1$ parameters $\theta$ and $D_1$ parameters $\theta_3$
\EndWhile
\State \textbf{EndWhile}
\State \textbf{return} $E_1$ parameters $\theta$
\EndWhile
\State \textbf{EndWhile}
\State Load 5 returned encoder parameter on five enocoders
\State Connect the outputs of 5 encoders and use the connected outputs as the input for the logistic regression model

\end{algorithmic}
\label{alg}
\end{algorithm}


\section{Experimental Validation}
\label{sec:experiment}

Our implementation code is publicly available.

\begin{figure*}[htb]
    \centering
    \includegraphics[width=10.8cm, height=8cm]{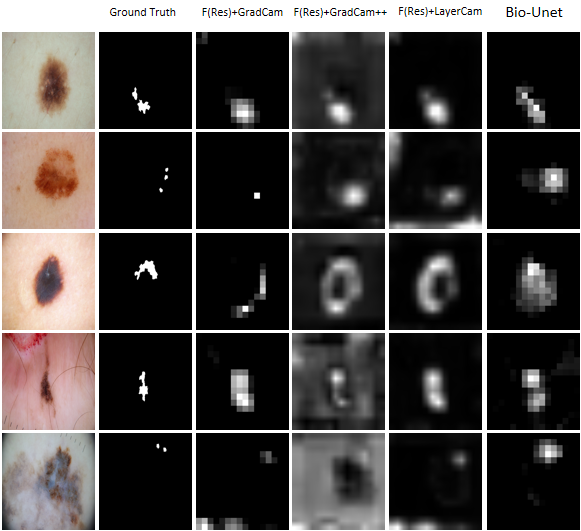}
    \caption{Localization performance for samples of dermatoscopic images: from top to the bottom, we have included a sample input image along with localization maps generated for globules, milia like cyst, negative network, pigment network , and streaks biomarkers indicators. From left to right, the input image, ground truth mask of the indicator, and masks generated by Grad-CAM, Grad-CAM++, LayerCAM, and Bio-Unet are visualized. The   cam-based feature maps are generated with a Resnet50 backbone,   trained   for classification.} 
    \label{fig:P2}
\end{figure*}

\subsection{Experimental Setup}

\paragraph{Datasets}
 We used the ISIC 2018 dataset~\cite{DBLP:journals/corr/abs-1902-03368,DBLP:journals/corr/abs-1803-10417}  to simulate our  semi-supervised learning framework. It is  a large collection of  dermatoscopic images of common pigmented skin lesions with several prediction tasks.  We used its Task 2 data as Dataset A and ISIC 2019 dataset~\cite{Tschandl2018-ft, Tschandl2018-xf, Combalia2019-te} as  Dataset B.

  \textbf{Dataset A:} 
 Task 2 of the ISIC 2018 dataset  poses a challenge for melanoma clinical indicator detection. The task is to detect the following five dermoscopic attributes that are melanoma indicators: pigment network, negative network, streaks, mila-like cysts and globules. For descriptions of these clinical indicator biomarkers, please refer to the ISIC 2018 documentation~\cite{DBLP:journals/corr/abs-1902-03368,DBLP:journals/corr/abs-1803-10417}. There are 2594 images in this task with binary labels for melanoma diagnosis. Table~\ref{tab1} shows the statistics for these indicators. As it can be guessed from our previous discussion, the dataset is sparse.  

\begin{center}
\begin{table}[h!]
\small
 \setlength\tabcolsep{2pt} 
 \centering
\begin{tabular}{@{}lccc@{}}
\toprule
                  & Nonempty  & empty  & Skin images total\\ \midrule
globules          & 603       & 1991   & 2594 \\
milia\_like\_cyst & 682       & 1912   & 2594 \\
negative network  & 190       & 2404   & 2594 \\
pigment network   & 1523      & 881    & 2594 \\     
streaks           & 100       & 2494   & 2594 \\   \bottomrule

\end{tabular}
\caption{\label{ISIC2018Task2} Comparison of the number of non-empty masks and the number of empty masks for each attribute}
\end{table}
\end{center}
\vspace{-5mm}
\label{tab1}

\textbf{Dataset B:} Task 3 in the ISIC 2019 dataset consists of 25331 images with only binary diagnosis labels. As it can be seen this dataset is much larger than Dataset A. However, the images are not annotated with the indicator biomarkers.

\paragraph{Baseline for Comparison:}

Bio-Unet was compared against using variants of CAM when applied on the classification subnetwork for generating heatmap explanations for each of the indicators. We included Layer-CAM, Grad-CAM, Grad-CAM++ in our experiments. Our goal is to demonstrate that despite having a good accuracy, the ability to classify images with high accuracy does not lead to human-interpretable explanations, irrespective of the particular algorithm that we use for generating heatmaps.


\paragraph{Evaluation metrics}
Our primary goal is localize the melanoma indicators on the input image. For this goal, we use the Continuous Dice metric~\cite{https://doi.org/10.48550/arxiv.1906.11031}. We computed the Continuous Dice metric between the generated mask for each indicator and the provided ground truth map. For melanoma diagnosis, we used classification  accuracy as the evaluation metric.


\paragraph{Implementation Details} For details about the algorithm implementation, optimizations hyperparameters, and the used hardware, please refer to the supplementary material.

\subsection{Performance Results}

After training the ResNet50 subnetwork and the Bio-UNet architecture for classification, we obtained $76\%$ and $82\%$ classification accuracy rates, respectively. We observe that both architecture have high melanoma diagnosis rates and using the additional subnetworks for segmentation and localization has  led to a significant diagnosis performance boost.
Table~\ref{ISIC2018Task22} presents results for localizing the five indicators when measured using the Continuous DICE metric. As it can be seen, utilizing Bio-UNet architecture along with our proposed training scheme  enhances localization results by 0.95\%, 2.33\%, 5.11\%,  -8.24\%, 9.01\%  over the standard architecture for the five indicators listed in Table ~\ref{tab1}, respectively. 
Each feature is critical, and Bio-UNet enables the encoder to better capture three features that are difficult for ResNet to detect. Every feature is vital for melanoma diagnosis, so when each feature can be localized uniformly, the accuracy of cancer diagnosis increases significantly. Table~\ref{tab2} presents results for localizing the five indicators when measured using the Continuous DICE metric. 

We find that when the model's decisions are close to what is expected by humans, the accuracy would increase. We conclude from our results that in order to make AI explainable, we may need to incorporate intermediate-level human-interpretable annotations in the deep learning end-to-end training pipelines and design DNN architectures that learn to perform a downstream task   using human-interpretable intermediate indicators.


\begin{center}
\begin{table}[!h]
\begin{adjustbox}{width=1\columnwidth}
\centering
 \setlength\tabcolsep{2pt} 
\begin{tabular}{@{}lcccccc@{}}
\toprule
                    & globules  & milia\_like\_cyst  & negative & pigment & streaks \\
                    \midrule
$f_{Res}(\cdot)$ + Grad-Cam            & 14.21     & 0.0           & 15.78   & 41.03   & 5.16    \\
$f_{Res}(\cdot)$ + Grad-Cam++          & 7.95      & 0.67          & 13.67   & 26.85   & 3.5     \\
$f_{Res}(\cdot)$ + Layer-Cam           & 14.53     & 0.0           & 15.89   & 40.67   & 5.83    \\
Bio-Unet                               & 15.16     & 2.33          & 20.89   & 32.79   & 14.17    \\     \bottomrule

\end{tabular}
\end{adjustbox}

\caption{\label{ISIC2018Task22} Evaluating the localization accuracy for the five clinical indicators. Continuous Dice coefficient in percentage is reported.}
\label{tab2}
\vspace{-5mm}
\end{table}

\end{center}

To provide an intuition behind the quality of results presented  in Table~\ref{tab2}, we have visualized samples of heatmaps that are generated using different methods in Figure \ref{fig:P2} for visual inspection. In this figure, we have selected an example image such that it is annotated with a corresponding indicator biomarker and presented the localization map that the AI pipelines generate. We can observe that the three CAM-based techniques generate features maps that are far larger than the ground truth mask and pretty much include the majority of the input mole. This means that the generated maps are not interpretable because they point to the whole mole which even a beginner knows that it should be the primary area of attention. In contrast, a close visualize comparison between columns two and six, demonstrates that our method generate binary feature maps that are quite similar to the ground truth, focusing on a subarea of the mole that in reality pertains to the clinical indicator. 
We also conclude that although the DICE metric is the predominant metric 
for segmentation, it is a sensitive metric when the semantic  classes in the images are imbalanced.

\subsection{Ablative Experiments}

\paragraph{Experiments on the importance of the subnetworks:}

\begin{figure}[ht]
    \centering
    \includegraphics[width=8cm, height=8cm]{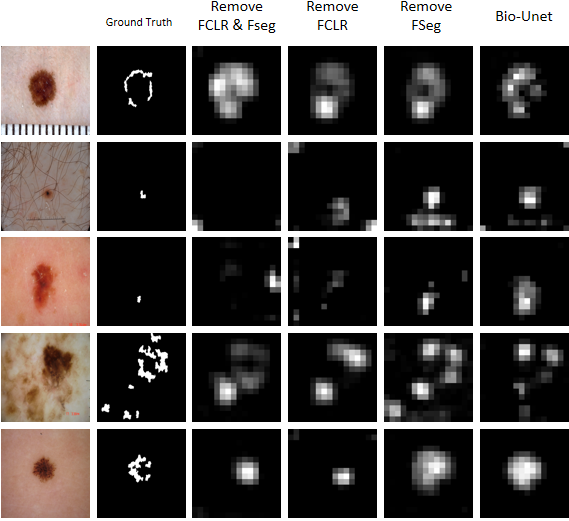}
    \caption{masks for the ablative study: from top to bottom are samples of globules, milia\_like\_cyst, negative network, pigment network, streaks. From left to right are input image,   ground truth mask, and mask generated by $f_{Res}(\cdot)$ , $f_{Res}(\cdot)$ + $f_{CLR}(\cdot)$  ,$f_{Res}(\cdot)$ + $f_{Seg}(\cdot)$ , and Bio-Unet, following Table~\ref{Tab3}.}
    \label{fig:P4}
\end{figure}

\begin{center}

\begin{table}[ht]
\setlength\tabcolsep{2pt} 
\centering
\begin{adjustbox}{width=1\columnwidth}
\begin{tabular}{@{}lccccccccccc@{}}
\toprule
                                                 & $f_{Seg}$        & $f_{CLR}$  & globules  & milia\_like\_cyst  & negative & pigment & streaks & Mel\\ 
                                                \midrule
$f_{Res}(\cdot)$                                 &                  &            & 14.21    & 0.0     & 15.78    & 41.03    & 5.16     & 0.76 \\
$f_{Res}(\cdot)$ + $f_{Seg}(\cdot)$              & \checkmark       &            & 15.21    & 0.5     & 17.78    & 39.58    & 3.83     & 0.76 \\
$f_{Res}(\cdot)$ + $f_{CLR}(\cdot)$              &                  & \checkmark & 15.11    & 1.0     & 17.44    & 39.53    & 12.67    & 0.80 \\
Bio-Unet                                         & \checkmark       & \checkmark & 15.16    & 2.33    & 20.89    & 32.79    & 14.17    & 0.82 \\     \bottomrule

\end{tabular}
\end{adjustbox}
\caption{\label{ISIC 2018 Task2} Localization performance results for the ablative study on the importance of subnetworks of Bio-Unet. }
\end{table}
 \label{Tab3}  
 \vspace{-10mm}
\end{center}

We conducted an ablation study to investigate the contribution of each component of Bio-UNet. Table~\ref{Tab3} presents the results of our ablative study. We observed that when the subnetwork $f_{CLR}(\cdot)$ was removed, the localization results for the "streaks" and "negative network" indicators were reduced. This observation was expected because "streaks" and "negative network" exist in only a very small number of samples and appear in a scattered and discontinuous manner in the input images. We concluded that self-supervised learning is extremely helpful for localizing infrequent indicators that appear in a scattered and discontinuous manner in the input images.

Figure \ref{fig:P4} presents samples of generated masks using our ablative experiment. The second row presents a case of the "milia-like cyst" indicator. It can be seen that this indicator corresponds to a single dot-like region in the ground truth. When we removed $f_{CLR}(\cdot)$, the encoder was unable to capture the accurate location of this region. However, when we added $f_{Seg}(\cdot)$ after $f_{CLR}(\cdot)$, the neural network was able to effectively remove irrelevant points and increase the importance of the region close to the ground truth. Therefore, we concluded that self-supervised learning can help provide more possible localization areas in some cases. From the last row of Figure \ref{fig:P4}, it can be seen that $f_{Seg}(\cdot)$ focuses on smaller regions, while $f_{CLR}(\cdot)$ provides a general region. When the two are combined, $f_{Seg}(\cdot)$ effectively refines the large but not confident region provided by $f_{CLR}(\cdot)$. We concluded that all the new aspects of our architecture are critical for improved performance.

\paragraph{Impact of repeating $f_{Seg}(\cdot)$ within one epoch:} 

\begin{center}
\begin{table}[!h]
\begin{adjustbox}{width=1\columnwidth}
\centering
 \setlength\tabcolsep{2pt} 
\begin{tabular}{@{}lcccccc@{}}
\toprule
                    & globules  & milia\_like\_cyst  & negative & pigment & streaks & Melanoma\\
                    \midrule
$f_{Res}(\cdot)$    & 14.21     & 0.0           & 15.78   & 41.03   & 5.16    & 0.76 \\
Bio-Unet            & 15.16     & 2.33          & 20.89   & 32.79   & 14.17   & 0.82\\
Repeat 2            & 14.89     & 2.67          & 21.22   & 37.92   & 15.50   & 0.80\\
Repeat 3            & 14.84     & 0.16          & 16.33   & 39.01   & 12.83   & 0.82\\
Repeat 4            & 13.68     & 1.00          & 19.00   & 43.19   & 7.93    & 0.81\\
Repeat 5            & 13.26     & 1.83          & 16.89   & 38.04   & 11.5    & 0.80\\     \bottomrule

\end{tabular}
\end{adjustbox}

\caption{\label{ISIC2018Task221} Evaluating the localization accuracy for the five clinical indicators. Continuous Dice Coefficient in percentage is reported.}
\label{tab5}
\vspace{-5mm}
\end{table}
\end{center}

\begin{figure}[h]
    \centering
    \includegraphics[width=8cm, height=7cm]{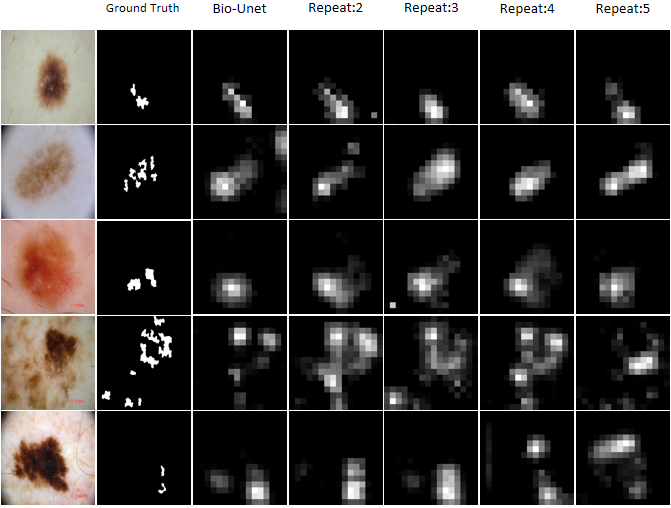}
    \caption{Impact of repeating $f_{Seg}(\cdot)$: the columns from left to right are ground truth, not repeated, repeated once, repeated twice,  repeated three times, repeated four times. From top to bottom are samples of globules, milia\_like\_cyst, negative network, pigment network, streaks.}
    \label{fig:P6}
\end{figure}

Finally, we conducted an experiment to investigate whether repeating $f_{Seg}(\cdot)$ within each epoch is helpful. Figure \ref{fig:P6} shows the results of repeating $f_{Seg}(\cdot)$ for all attributes. We can observe the second and third columns (last row) of the "streaks" indicator, which correspond to no repetition and repetition once, respectively. We observed that when we repeated $f_{Seg}(\cdot)$ once in the loop, the resulting localization mask successfully focused on the two discontinuous parts and abandoned the wrong attention area on the left, compared to the Bio-UNet without repetition. Additionally, from Table~\ref{tab5}, we can see that four out of five features increased, while the diagnostic rate for melanoma decreased. 
However, it can be seen from Table~\ref{tab5} that when repeating for the fourth time, the localization score of "pigment network" reached its highest value, but the localization score of "streaks" decreased. Therefore, I believe that each feature should use a different number of repetitions of $f_{Seg}(\cdot)$.  But for consistency in the paper, we ultimately used only one repetition of $f_{Seg}(\cdot)$ for all attributes.

\paragraph{Five encoders instead of multi-task learning}

\begin{center}
\begin{table}[!h]
\begin{adjustbox}{width=1\columnwidth}
\centering
 \setlength\tabcolsep{2pt} 
\begin{tabular}{@{}lcccccc@{}}
\toprule
                            & globules  & milia\_like\_cyst  & negative & pigment & streaks & Melanoma\\
                            \midrule
$f_{Res}(\cdot)$            & 14.21     & 0.0           & 15.78   & 41.03   & 5.16    & 0.76 \\
Bio-Unet                    & 15.16     & 2.33          & 20.89   & 32.79   & 14.17   & 0.82\\
Bio-Unet-(Two-tasks)         & 14.16     & 1.16          & 20.22   & 37.88   & 14.33   & 0.84\\
Bio-Unet-(Five-tasks)        & 15.89     & 0.5           & 19.44   & 15.64   & 10.67   & 0.82\\
Bio-Unet-(Six-tasks)         & 14.68     & 1.5           & 19.33   & 23.21   & 11.83   & 0.83\\     \bottomrule

\end{tabular}
\end{adjustbox}

\caption{\label{ISIC2018Task222} Evaluate the localization accuracy of five clinical indicators in different mlutitask learning. Continuous Dice Coefficient in percentage is reported.}
\label{tab6}
\vspace{-5mm}
\end{table}
\end{center}

Table \ref{tab6} represents different multi-task learning approaches. Bio-Unet-(Two-tasks) denotes that the encoder predicts both one intermediate feature and Melnoma label, while Bio-Unet-(Five-tasks) denotes that the encoder predicts all five intermediate features simultaneously. Bio-Unet-(Six-tasks) denotes that the encoder predicts all five intermediate features as well as the Melnoma label. We can see that adding the Melnoma label significantly improves the AUC, reaching a maximum of 84\%, and the localization scores  for "pigment" and "streaks" also increase. This shows that multi-task learning with Melnoma plays a significant role in improving performance. However, as the number of tasks increases to five or six, the continual dice coefficient for most attributes decreases, although the AUC remains sufficiently high.

I believe that performing $f_{Seg}(\cdot)$ for each individual attribute may conflict with the multi-task learning of Five or Six tasks. Therefore, I chose to use five encoders for each intermediate feature, which resulted in better cDC.


\section{Conclusions}
\label{sec:conclusion}
We developed an architecture for explainable diagnosis of melanoma using skin lesion images. Our architecture is designed to localize melanoma clinical indicators spatially and use them to predict the diagnosis label. As a result, it performs the task similar to a clinician, leading to interpretable decisions. We benefited from contrastive learning and self-supervised learning to address the challenge of annotated data scarcity for our task that requires coarse annotations with respect to clinical indicators. Experimental results demonstrate that our model is able to generate localization masks for identifying clinical biomarkers and generates more plausible explanations compared to existing classification architectures. Future works include extensions to  learning settings with distributed data~\cite{rostami2018multi}. 

{\small
\bibliographystyle{ieee_fullname}
\bibliography{egbib}
}

\clearpage

\appendix

\section{Implementation Details}

\begin{figure*}[htb]
    \centering
    \includegraphics[width=18cm, height=10cm]{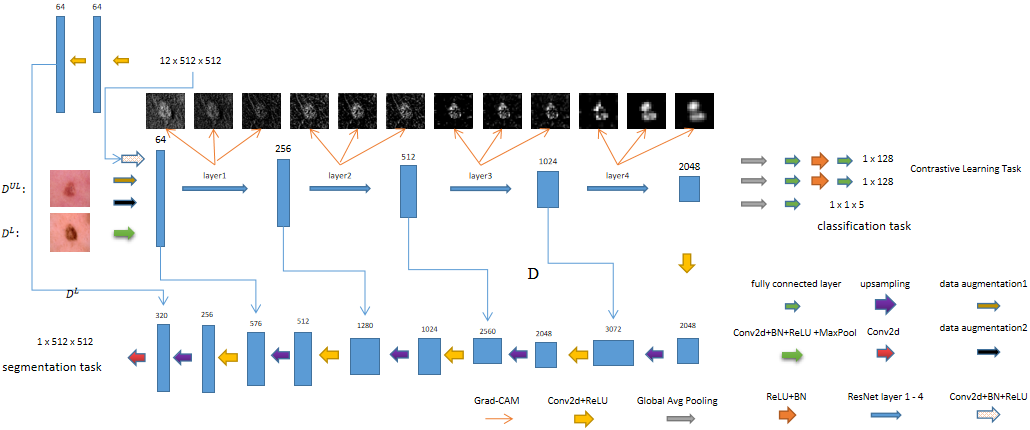}
    \caption{Proposed architecture for explainable diagnosis of melanoma: the architecture is trained to simultaneously classify skin lesion pictures using a CNN classifier and learn to localize melanoma clinical indicators on the input image using a U-Net-based segmentation network. The classification branch receive its input from the segmentation path to enforce classifying images based on clinical indicators.}
    \label{fig:P1a}
\end{figure*}

We provide details of our experimental implementation.

\paragraph{Hardware and Optimizer:}
The entire framework is implemented using PyTorch and trained on four NVIDIA RTX 2080Ti GPUs with 11GB of memory. We use ResNet50 with pretrained weights on ImageNet as the encoder for our architecture Bio-Unet. The Adam optimizer with one set of hyperparameters (lr = $1e^{-4}$, weight decay = $1e^{-4}$) is used for all tasks during the training stage.

\subsection{Optimization Implementation Details}

\paragraph{Preprocessing:}
We start by training the subnetwork $f_{CLR}(\cdot)$. Dataset B serves as the input for $f_{CLR}(\cdot)$. Each image is first resizes to $512 \times 512$, then the image  is normalized to have a zero mean and unit variance. Finally, data augmentation is carried out to improve the model generalization. Operations for data augmentation include random rotation, cropping, brightness, contrast, saturation, and flipping. Each mini-batch contains one positive example and 22 negative examples, with the batch size set to 24. The temperature parameter $\tau$ is set to 0.5.

We then perform the classification task using all of dataset A's data, each resized to $512 \times 512$, and normalize images with zero mean and unit variance without performing any data augmentation. Due to the large size of the images and memory cap, the batch size is set to 16.

\paragraph{Segmentation training:}
Once the classification task is complete, we use   Grad-CAM   to create an attention heatmap with a bottleneck block and the desired target attribute index as input. We then use this method to replace the bottlenecks one at a time until every bottleneck has been tried. After gathering all heatmaps, we input those heatmaps into the $f_{Seg}(\cdot)$ subnetwork for training it as the reconstruction function. We used  a batch size of 8.

After the network training is complete, we create heatmaps using   Grad-CAM as $f_{Grad\_CAM}(b_i,j)$ , where $b_i\in E_1(\cdot)$, using the bottleneck block $b_i$ in the encoder $E_1(\cdot)$ loaded with the optimum checkpoint parameters and the index of the selected attribute as input. We would select the heatmap of the last block as the final localization mask for the biomarker indicators.

\begin{figure*}[htb]
    \centering
    \includegraphics[width=15cm, height=18cm]{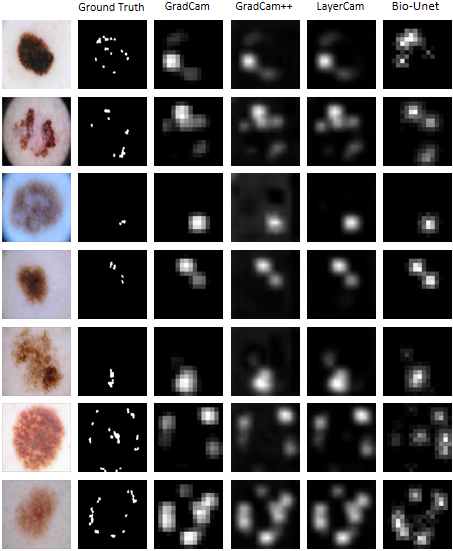}
    \caption{More examples of feature globules; Examples showing that $f_{Seg}$ can make the framework focus more on small important regions rather than large ones }
    \label{fig:P2a}
\end{figure*}

\begin{figure*}[htb]
    \centering
    \includegraphics[width=15cm, height=18cm]{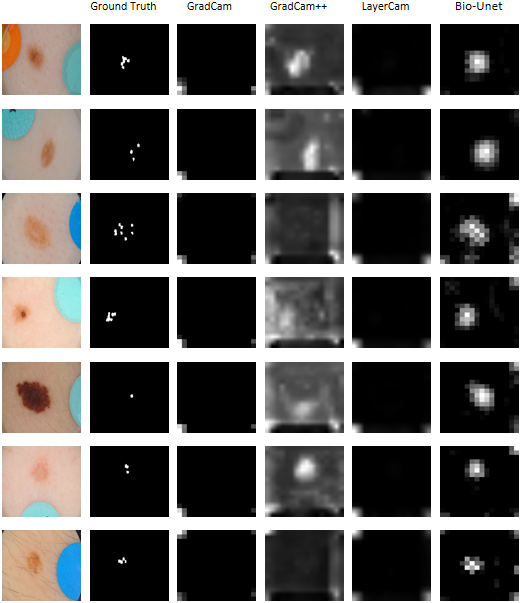}
    \caption{More examples of feature Milia like Cyst; Example showing that $f_{CLR}$ can help framewrok find large region of small scatter points and $f_{Seg}$ can make the framework focus more on small important regions rather than large ones }
    \label{fig:P3a}
\end{figure*}

\begin{figure*}[htb]
    \centering
    \includegraphics[width=15cm, height=18cm]{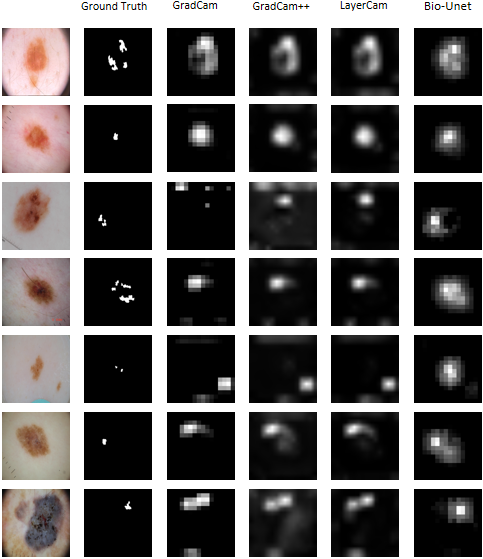}
    \caption{More examples of feature negative network; Example showing that $f_{Seg}$ can make the framework focus more on small important regions rather than large ones }
    \label{fig:P4a}
\end{figure*}

\begin{figure*}[htb]
    \centering
    \includegraphics[width=15cm, height=18cm]{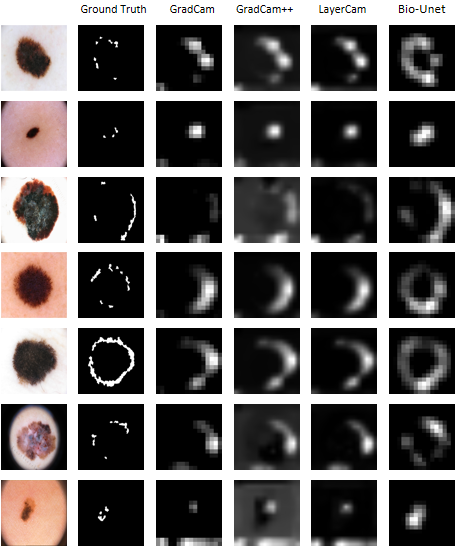}
    \caption{More examples of feature streaks; Example showing that $f_{CLR}$ can help framewrok find large region of small scatter points and $f_{Seg}$ can make the framework focus more on small important regions rather than large ones }
    \label{fig:P5a}
\end{figure*}


\end{document}